\documentclass[letterpaper, 10 pt, conference]{ieeeconf}  

\usepackage{amsmath,amsfonts}
\usepackage{makecell}
\usepackage{algorithm}
\usepackage{algorithmic}
\usepackage{array}
\usepackage[caption=false,font=normalsize,labelfont=sf,textfont=sf]{subfig}
\usepackage{textcomp}
\usepackage{stfloats}
\usepackage{pifont}
\usepackage{cite}
\makeatletter
\let\NAT@parse\undefined
\makeatother
\usepackage{lineno,hyperref}
\usepackage{booktabs} 

\hypersetup{
	colorlinks=true, 
	linkcolor=blue,  
	citecolor=blue, 
	urlcolor=red     
}
\usepackage{url}
\usepackage{verbatim}
\usepackage{graphicx}
\usepackage{xcolor}

\def\BibTeX{{\rm B\kern-.05em{\sc i\kern-.025em b}\kern-.08em
		T\kern-.1667em\lower.7ex\hbox{E}\kern-.125emX}}
\usepackage{balance}

\title{\LARGE \bf
Real-Time Whole-Body Safe Motion Generation for Multi-Segment Tendon-Driven Continuum Robots}

\author{Fangju Yang$^{1,2}$,  Siyi Ma$^{1{\dagger}}$, Tonghao Guan$^{3{\dagger}}$, Tingcong Liu$^{1}$, Hang Yang$^{1}$, Zhengqiang Zhang$^{2}$\\Jian S. Dai$^{4}$, Ke Wu$^{1}$ 
}

\IEEEoverridecommandlockouts                              

\begin{document}

\maketitle
\thispagestyle{empty}
\pagestyle{empty}


\begin{abstract}
Real-time motion generation for tendon-driven continuum robots requires
accurate modeling of nonuniform bending and whole-body collision avoidance.
This paper presents a unified actuation-space framework for planar
multi-segment tendon-driven continuum robots. An energy-based variable-curvature
model captures spatially varying tendon spacing and bending stiffness and
provides analytical point Jacobians for differential inverse kinematics and
safety monitoring. A multipoint CBF-QP enforces backbone clearance under
obstacle motion and actuation-velocity bounds, while its decision dimension
depends only on the number of independently actuated segments. The model
closely agrees with GVS references, with a maximum curvature error of
$5.223\times10^{-2}\,\mathrm{m}^{-1}$. Over 100 MuJoCo trials, the proposed
method achieves collision-free success rates of 96\% and 100\% in static and
dynamic scenarios, respectively, compared with approximately 60\% and 80\%
without CBF constraints. Hole-traversal tests further demonstrate safe motion
in constrained environments. With 600 backbone monitoring points, the mean
control-step time is $6.66\,\mathrm{ms}$, demonstrating real-time whole-body
safe motion generation.
\end{abstract}


\section{INTRODUCTION}
Continuum robots exploit continuously deformable backbones to achieve high conformability and inherent compliance in narrow, confined, and cluttered environments, making them well suited for minimally invasive intervention, in-situ inspection, and maintenance tasks~\cite{burgner2015continuum,russo2023continuum}. Among them, tendon-driven continuum robots (TDCRs) place the actuators outside the flexible body while enabling independent actuation of multiple segments and retaining a slender morphology~\cite{rao2021modeling,rucker2011statics}. Real-time motion control in complex environments, however, remains challenging for two main reasons. First, structural properties such as tendon spacing, cross-sectional geometry, and bending stiffness may vary along the backbone, leading to spatially nonuniform intra-segment curvature that cannot be accurately captured by conventional constant-curvature models~\cite{mahl2014variable,renda2020geometric}. Second, collision risks are distributed along the entire continuous body rather than confined to the tip, and tip-only planning or control is therefore insufficient to ensure whole-body safety~\cite{ataka2016safer,li2024active}. Consequently, simultaneously achieving accurate nonuniform variable-curvature modeling, real-time computation, and whole-body safety control remains a central challenge for TDCRs operating in confined and dynamic environments.

Recent studies on efficient continuum-robot kinematics have largely focused on the tradeoff between computational efficiency and configuration accuracy. Piecewise constant curvature (PCC) models provide compact analytical mappings with low computational cost and have therefore been widely adopted for real-time kinematics and control~\cite{jones2006kinematics,webster2010design}. Their accuracy, however, relies on the assumption of approximately constant curvature within each segment, and spatial variations in structural parameters can significantly degrade configuration prediction~\cite{camarillo2008mechanics,mahl2014variable}. In contrast, continuum-mechanics models based on Euler--Bernoulli beam theory or Cosserat rod theory can capture more general elastic deformation and tendon-routing effects~\cite{rucker2011statics,janabi2021cosserat}, but they typically require the solution of spatial boundary-value problems and thus incur higher online computational cost~\cite{till2019realtime,rao2021modeling}. To relax the constant-curvature assumption while retaining computational efficiency, polynomial-curvature representations and variable-strain formulations have been investigated~\cite{mahl2014variable,renda2020geometric}, while analytical differential kinematics and actuation-space energy models have further improved real-time modeling capability~\cite{ba2024analytical,Wu2026LASEM}. Despite these advances, multi-segment TDCRs with spatially nonuniform structural parameters still lack a compact framework that consistently provides real-time forward kinematics, analytical differential mappings at arbitrary backbone locations, and inverse kinematics under a unified actuation representation.

Whole-body collision avoidance for continuum robots has been studied through motion planning, reactive control, and optimization-based methods~\cite{rao2024contact,ataka2016safer,hachen2025nonlinear}. Search- and sampling-based planners can generate collision-free configurations in cluttered or confined environments, but their computation is typically performed offline or within a receding-horizon framework, which limits their ability to impose instantaneous safety constraints at every control cycle~\cite{rao2024contact,gao2024path}. Reactive approaches instead exploit backbone-state or local-motion information to modify the robot configuration online and can achieve high-rate obstacle avoidance in dynamic environments~\cite{ataka2016safer,li2024active}. However, these methods generally lack explicit forward-invariance guarantees for the safe set. Optimization-based approaches, such as nonlinear model predictive control, can incorporate shape, collision, and actuation constraints~\cite{hachen2025nonlinear}, but their online complexity grows with the prediction horizon, model dimension, and number of constraints, limiting high-frequency execution.

Control barrier functions (CBFs) provide an alternative formulation in which safety requirements are expressed as affine constraints on the control input, enabling forward invariance of the safe set while minimally modifying a nominal command through quadratic programming~\cite{ames2017control}. More recently, Wong et al.\ developed a closed-form CLF-CBF controller for continuum soft robots by aggregating dense collision constraints through Log-Sum-Exp, thereby eliminating online QP solution and enabling real-time whole-body obstacle avoidance~\cite{wong2026closedform}. This result demonstrates the feasibility of high-rate whole-body safety control, but the formulation relies on a piecewise constant-strain model and does not jointly account for spatially varying backbone mechanics, dynamic-obstacle velocity terms, or actuation-velocity bounds. Consequently, a real-time safety framework that remains consistent with a nonuniform variable-curvature kinematic model while retaining explicit whole-body safety constraints is still needed.

This paper develops a unified actuation-space framework for variable-curvature kinematics and whole-body safe motion generation. The main contributions are as follows:
\begin{enumerate}
    \item \textbf{Closed-form modeling:} An energy-based variable-curvature model captures nonuniform tendon spacing and bending stiffness, providing closed-form kinematics and analytical backbone Jacobians, with maximum curvature errors below \(5.3\times10^{-2}\,\mathrm{m}^{-1}\) against GVS references.
    \item \textbf{Whole-body safe motion generation:} A multi-point CBF-QP framework enables real-time whole-body safe motion generation, achieving collision-free success rates of 96\% and 100\% in static and dynamic MuJoCo scenarios, respectively, over 100 trials, with further validation in hole-traversal tasks.
    \item \textbf{Computational efficiency and scalability:} The QP decision dimension depends only on the number of independently actuated segments, enabling dense safety monitoring with a mean control-step time of $6.66\,\mathrm{ms}$ for 600 monitoring points.
\end{enumerate}

\section{Variable-Curvature Kinematic Modeling in Actuation Space}
\begin{figure*}[!t]
    \centering \includegraphics[width=0.92\textwidth]{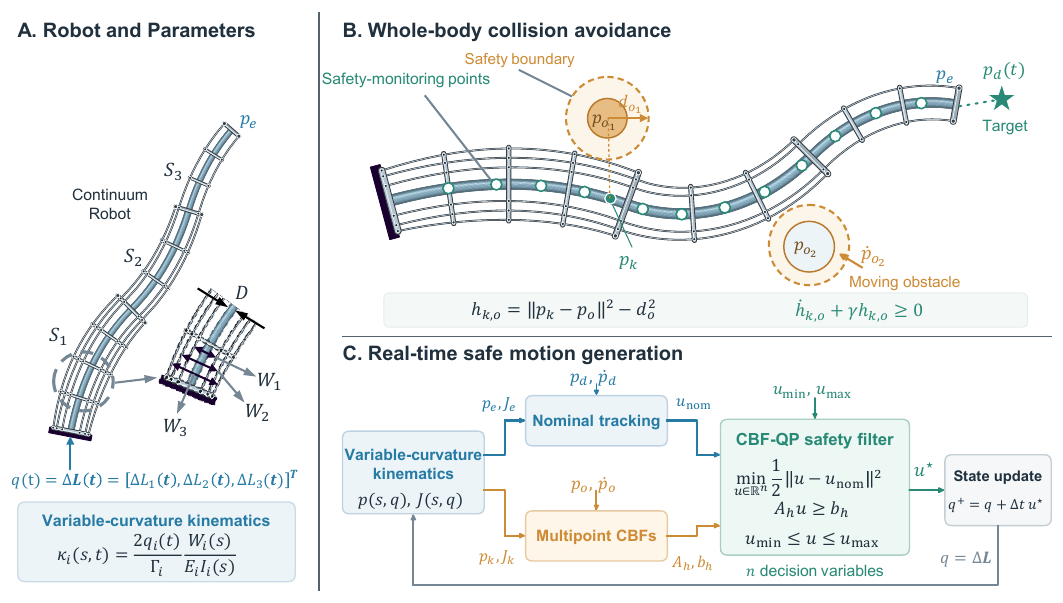}
    \caption{Real-time whole-body safe motion generation for continuum robots. (A) Robot and kinematics. (B) Whole-body collision avoidance. (C) Real-time CBF-QP control framework.}
    \label{fig:method_overview}
\end{figure*}
This paper considers the $n$-segment planar tendon-driven continuum robot (TDCR) shown in Fig.~\ref{fig:method_overview} and develops a general variable-curvature kinematic model in actuation space. The robot comprises $n$ serially connected flexible segments, where the $i$-th segment is denoted by $\mathcal{S}_i$ and has length $L_i$, yielding the total robot length $L=\sum_{i=1}^{n}L_i$. The arc-length coordinate $s\in[0,L]$ specifies a material point along the backbone. To capture the spatially varying curvature induced by structural nonuniformity, each segment may have arc-length-dependent geometric and mechanical parameters, including the cross-sectional diameter $D_i(s)$, second moment of area $I_i(s)$, and tendon spacing $W_i(s)$, while $E_i$ denotes the Young's modulus of the $i$th segment. Each flexible segment is independently actuated by a pair of inextensible tendons symmetrically routed along opposite sides of the backbone. For the $i$th segment, let the differential tendon-length variation be $\Delta L_i(t)$, and define the actuation vector $\Delta\boldsymbol{L}(t)=[\Delta L_1(t),\,\Delta L_2(t),\,\ldots,\,\Delta L_n(t)]^{\mathrm T}$. Here, $\Delta L_i(t)$ describes the independent displacement actuation of the $i$th TDCR segment and determines its bending configuration through the local geometric constraints and elastic deformation. Let $\theta(s,t)$ denote the local tangent angle of the backbone at arc-length position $s$ with respect to the global frame. The local curvature of the $i$th segment is $\kappa_i(s,t)=\theta_i'(s,t)$, and the backbone position in the global frame is $\boldsymbol{p}(s,t)=[x(s,t),\,y(s,t)]^{\mathrm T}$. Based on this parameterization, the mapping from the independent multi-segment displacement inputs to the continuous curvature field is established, followed by the forward, inverse, and differential kinematic models from actuation space to Cartesian space at arbitrary backbone positions. Throughout this paper, $\dot{(\cdot)}=\partial(\cdot)/\partial t$, $\ddot{(\cdot)}=\partial^2(\cdot)/\partial t^2$, 
$(\cdot)'=\partial(\cdot)/\partial s$, and $(\cdot)''=\partial^2(\cdot)/\partial s^2$.

\subsection{Geometric Relations and Energy Functional}
Consider an inextensible robot backbone whose planar configuration is described by the local tangent angle $\theta(s,t)$. The backbone position in the global frame is given by
\begin{equation}
\small
\boldsymbol{p}(s,t)
=
\begin{bmatrix}
x(s,t)\\
y(s,t)
\end{bmatrix}
=
\int_{0}^{s}
\begin{bmatrix}
\cos\theta(\xi,t)\\
\sin\theta(\xi,t)
\end{bmatrix}
\mathrm{d}\xi .
\label{eq:backbone_geometry}
\end{equation}
To characterize the piecewise geometry of the multi-segment flexible backbone, define the cumulative arc-length coordinates as
\begin{equation}
\small
S_0=0,\quad
S_i=\sum_{j=1}^{i}L_j,\quad
S_n=L.
\end{equation}
Accordingly, the $i$-th flexible segment occupies the arc-length interval $s\in[S_{i-1},S_i]$. At any time $t$, its local curvature satisfies
$\kappa_i(s,t)=\theta_i'(s,t)$, $s\in[S_{i-1},S_i]$.

Following the geometric relation in~\cite{Wu2026LASEM}, for two inextensible tendons symmetrically routed along both sides of the backbone, the tendon-length variation can be expressed in terms of the backbone curvature and local tendon spacing as
\begin{equation}
\small
\Delta L_i(t)
=
\frac{1}{2}
\int_{S_{i-1}}^{S_i}
W_i(\xi)\theta_i'(\xi,t)\,\mathrm{d}\xi ,\quad i=1,\ldots,n.
\label{eq:cable_length_constraint}
\end{equation}
This relation imposes a geometric constraint between the differential tendon-length variation and the local continuous curvature field of the $i$th segment while allowing the tendon spacing $W_i(s)$ to vary continuously along the arc length.

To further characterize the elastic deformation associated with the curvature field, the flexible backbone is modeled using Euler--Bernoulli beam theory. The bending strain energy of the $i$-th segment is
\begin{equation*}
\small
U_i
=
\frac{1}{2}
\int_{S_{i-1}}^{S_i}
E_i I_i(\xi)
\left[\theta_i'(\xi,t)\right]^2
\mathrm{d}\xi ,
\end{equation*}
and the total bending strain energy of the robot backbone is therefore
\begin{equation}
\small
U
=
\frac{1}{2}
\sum_{i=1}^{n}
\int_{S_{i-1}}^{S_i}
E_i I_i(\xi)
\left[\theta_i'(\xi,t)\right]^2
\mathrm{d}\xi .
\label{eq:bending_energy}
\end{equation}

Following~\cite{yang2026lightweightC}, 
the equilibrium configuration under a prescribed actuation input is obtained from the following constrained variational problem:
\begin{equation}
\small
\begin{aligned}
\theta^{*}(s,t)
&=
\arg\min_{\theta(s,t)}
U[\theta],
\\
&\text{s.t.}\quad
\eqref{eq:cable_length_constraint}.
\end{aligned}
\label{eq:functional_minimization}
\end{equation}
Thus, determining the multi-segment continuous curvature field $\theta_i'(s,t)$ reduces to a variational problem for the potential-energy functional subject to the differential tendon-length constraints.

\subsection{Variable-Curvature Forward Kinematic Mapping}
To determine the continuous curvature field for prescribed tendon-length variations, Eq.~\eqref{eq:cable_length_constraint} is imposed as the geometric constraint for each flexible segment, and a constrained variational problem is formulated using the principle of minimum potential energy. For the $i$th flexible segment, the tendon-length variation satisfies
\begin{equation*}
\small
\Delta L_i(t)
-
\frac{1}{2}
\int_{S_{i-1}}^{S_i}
W_i(\xi)\theta_i'(\xi,t)
\,\mathrm{d}\xi
=
0,
\quad
i=1,\ldots,n.
\end{equation*}

To solve the constrained variational problem in~\eqref{eq:functional_minimization}, introduce a time-varying Lagrange multiplier $\lambda_i(t)$ for the differential tendon-length constraint of each segment and construct the augmented functional
\begin{equation}
\small
\begin{aligned}
\mathcal{H}[\theta,\lambda]
=&\,
U[\theta]
+\!\!
\sum_{i=1}^{n}
\lambda_i(t)
\Bigg[
\Delta L_i(t)
-
\frac{1}{2}\!
\int_{S_{i-1}}^{S_i}\!\!\!\!
W_i(\xi)\theta_i'(\xi,t)
\,\mathrm{d}\xi
\Bigg],
\end{aligned}
\label{eq:augmented_functional}
\end{equation}
where $\lambda_i(t)$ is the Lagrange multiplier associated with the differential tendon-length constraint of the $i$th segment. The equilibrium configuration must satisfy the stationarity conditions with respect to both $\theta_i(s,t)$ and $\lambda_i(t)$, i.e.,
\begin{equation*}
\small
\delta_{\theta}\mathcal{H}=0,
\quad
\delta_{\lambda}\mathcal{H}=0.
\end{equation*}

Variation with respect to $\lambda_i(t)$ directly recovers the differential tendon-length constraint in~\eqref{eq:cable_length_constraint}. Further perturbing $\theta_i(s,t)$ by an arbitrary admissible variation $\delta\theta_i(s,t)$ gives
\begin{equation}
\small
\delta_{\theta}\mathcal{H}
=
\sum_{i=1}^{n}\!\!
\int_{S_{i-1}}^{S_i}\!\!\!
\left[
E_iI_i(\xi)\theta_i'(\xi,t)
\!-\!
\frac{1}{2}\lambda_i(t)W_i(\xi)
\right]
\delta\theta_i'(\xi,t)
\,\mathrm{d}\xi .
\label{eq:first_variation}
\end{equation}
Integrating the term associated with each segment in~\eqref{eq:first_variation} by parts gives
\begin{equation}
\small
\begin{aligned}
\delta_{\theta}\mathcal{H}
=&\!
-\!\!\sum_{i=1}^{n}\!\!
\int_{S_{i-1}}^{S_i}\!\!
\frac{\mathrm{d}}{\mathrm{d}\xi}\!\!
\left[
E_iI_i(\xi)\theta_i'(\xi,t)
-
\frac{1}{2}\lambda_i(t)W_i(\xi)
\right]
\delta\theta_i(\xi,t)
\,\mathrm{d}\xi
\\
&
+
\sum_{i=1}^{n}
\left[
\left(
E_iI_i(\xi)\theta_i'(\xi,t)
-
\frac{1}{2}\lambda_i(t)W_i(\xi)
\right)
\delta\theta_i(\xi,t)
\right]_{S_{i-1}}^{S_i}.
\end{aligned}
\label{eq:variation_integration_by_parts}
\end{equation}

Because the perturbation $\delta\theta_i(\xi,t)$ is arbitrary within each segment, the variational equilibrium equation for the $i$th flexible segment is
\begin{equation}
\small
\frac{\mathrm{d}}{\mathrm{d}s}
\left[
E_iI_i(s)\theta_i'(s,t)
-
\frac{1}{2}\lambda_i(t)W_i(s)
\right]
=
0,
\quad
s\in[S_{i-1},S_i].
\label{eq:EL_general}
\end{equation}
Geometric continuity of the multi-segment backbone further requires the tangent angles of adjacent segments to satisfy
\begin{equation*}
\small
\theta_i(S_i,t)
=
\theta_{i+1}(S_i,t),
\quad
i=1,\ldots,n-1 .
\label{eq:theta_continuity}
\end{equation*}
The boundary terms in~\eqref{eq:variation_integration_by_parts}, together with continuity of the tangent-angle variation at each junction, yield the interface condition between adjacent flexible segments
\begin{equation}
\small
\begin{aligned}
&\left[
E_iI_i(s)\theta_i'(s,t)
-
\frac{1}{2}\lambda_i(t)W_i(s)
\right]_{s=S_i^-}
\\
&\quad =
\left[
E_{i+1}I_{i+1}(s)\theta_{i+1}'(s,t)
-
\frac{1}{2}\lambda_{i+1}(t)W_{i+1}(s)
\right]_{s=S_i^+}.
\end{aligned}
\label{eq:interface_condition}
\end{equation}
Together with the fixed-base condition $\theta(0,t)=0$, the absence of an external moment at the robot tip gives the natural boundary condition
\begin{equation}
\small
E_nI_n(L)\theta_n'(L,t)
-
\frac{1}{2}\lambda_n(t)W_n(L)
=
0.
\label{eq:natural_boundary}
\end{equation}

Integrating the segment-wise variational equilibrium equation gives
\begin{equation*}
\small
E_iI_i(s)\theta_i'(s,t)
-
\frac{1}{2}\lambda_i(t)W_i(s)
=
C_i(t).
\end{equation*}
Combining this result with the interface condition~\eqref{eq:interface_condition} yields
\begin{equation*}
\small
C_i(t)=C_{i+1}(t),
\quad
i=1,\ldots,n-1,
\end{equation*}
Applying the tip natural boundary condition~\eqref{eq:natural_boundary} further gives
\begin{equation*}
\small
C_1(t)=C_2(t)=\cdots=C_n(t)=0.
\end{equation*}
Therefore, the local curvature of the $i$th flexible segment satisfies
\begin{equation}
\small
\kappa_i(s,t)
=
\theta_i'(s,t)
=
\frac{\lambda_i(t)W_i(s)}
{2E_iI_i(s)},
\quad
s\in[S_{i-1},S_i].
\label{eq:variable_curvature_multiplier}
\end{equation}
Substituting~\eqref{eq:variable_curvature_multiplier} into the tendon-length geometric constraint~\eqref{eq:cable_length_constraint} for the $i$th segment gives
\begin{equation*}
\small
\Delta L_i(t)
=
\frac{\lambda_i(t)}{4}
\int_{S_{i-1}}^{S_i}
\frac{W_i^2(\xi)}
{E_iI_i(\xi)}
\,\mathrm{d}\xi .
\end{equation*}
Define the actuation-compliance coefficient of the $i$th segment as
\begin{equation*}
\small
\Gamma_i
=
\int_{S_{i-1}}^{S_i}
\frac{W_i^2(\xi)}
{E_iI_i(\xi)}
\,\mathrm{d}\xi ,
\label{eq:gamma_definition}
\end{equation*}
The Lagrange multiplier is then determined directly from the prescribed differential tendon-length variation as
\begin{equation}
\small
\lambda_i(t)
=
\frac{4\Delta L_i(t)}
{\Gamma_i}.
\label{eq:multiplier_length_relation}
\end{equation}
Substituting~\eqref{eq:multiplier_length_relation} into~\eqref{eq:variable_curvature_multiplier} gives the local curvature field in terms of the tendon-length variation $\Delta L_i(t)$
\begin{equation}
\small
\kappa_i(s,t)
=
\frac{2\Delta L_i(t)}
{\Gamma_i}
\frac{W_i(s)}
{E_iI_i(s)},
s\in[S_{i-1},S_i].
\label{eq:variable_curvature_length}
\end{equation}
Thus, for prescribed differential tendon-length variations, the local curvature is jointly determined by the actuation input $\Delta L_i(t)$ and the spatial distributions of the local tendon spacing $W_i(s)$ and bending stiffness $E_iI_i(s)$.

To describe the relative rotation induced by independent actuation within the $i$th segment, define the cumulative bending-compliance function
\begin{equation*}
\small
\Xi_i(s)
=
\int_{S_{i-1}}^{s}
\frac{W_i(\xi)}
{E_iI_i(\xi)}
\,\mathrm{d}\xi ,
\quad
s\in[S_{i-1},S_i].
\label{eq:beta_definition}
\end{equation*}
Integrating~\eqref{eq:variable_curvature_length} along the arc length, the $i$th segment generates the following relative rotation from its start $S_{i-1}$ to any position $s$:
\begin{equation*}
\small
\vartheta_i(s,t)
=
\frac{2\Delta L_i(t)}
{\Gamma_i}
\Xi_i(s).
\label{eq:segment_relative_angle_length}
\end{equation*}
Therefore, for any $s\in[S_{i-1},S_i]$, the backbone tangent angle in the global frame comprises the full rotations of all preceding segments and the local rotation of the current segment, i.e.,
\begin{equation*}
\small
\theta_i(s,t)
=
\sum_{j=1}^{i-1}
\vartheta_j(S_j,t)
+
\vartheta_i(s,t).
\end{equation*}
The global tangent angle of the displacement-actuated backbone is therefore
\begin{equation}
\small
\theta_i(s,t)
=
2
\left[
\sum_{j=1}^{i-1}
\frac{\Delta L_j(t)}
{\Gamma_j}
\Xi_j(S_j)
+
\frac{\Delta L_i(t)}
{\Gamma_i}
\Xi_i(s)
\right].
\label{eq:theta_length_global}
\end{equation}

Finally, substituting~\eqref{eq:theta_length_global} into the backbone geometry~\eqref{eq:backbone_geometry} yields the forward kinematic mapping from the multidimensional displacement-actuation space to Cartesian space. 


\subsection{Residual-Based Variable-Curvature Inverse Kinematics}
Building on the forward kinematic model, consider the inverse kinematic problem of recovering the independent multi-segment actuation inputs from a desired tip trajectory. To address the nonlinear mapping between the tip configuration and the actuation vector $\Delta\boldsymbol{L}(t)$, a differential kinematic relation between actuation space and Cartesian space is established.

From~\eqref{eq:theta_length_global}, the influence of the $j$th sectional input $\Delta L_j(t)$ on the backbone tangent angle is
\begin{equation}
\small
\frac{\partial\theta(s,t)}
{\partial\Delta L_j(t)}
=
\begin{cases}
0,
& 0\le s<S_{j-1},
\\[2mm]
\dfrac{2\Xi_j(s)}{\Gamma_j},
& S_{j-1}\le s\le S_j,
\\[3mm]
\dfrac{2\Xi_j(S_j)}{\Gamma_j},
& S_j<s\le L.
\end{cases}
\label{eq:angle_actuation_influence}
\end{equation}
The $j$-th sectional actuation does not affect the tangent angle of the preceding backbone, but produces a cumulative orientation change in the current and all downstream segmens.

For compactness, define the angular influence function associated with the $j$th actuation input as
\begin{equation*}
\small
\phi_j(s)
=
\frac{\partial\theta(s,t)}
{\partial\Delta L_j(t)},
\quad
j=1,\ldots,n,
\label{eq:angle_influence_function}
\end{equation*}
and construct the actuation influence vector
\begin{equation}
\small
\boldsymbol{\phi}(s)
=
\begin{bmatrix}
\phi_1(s),\phi_2(s),\cdots,\phi_n(s)
\end{bmatrix}^{\mathrm T}.
\label{eq:angle_influence_vector}
\end{equation}
The tangent angle at any backbone position $s$ can then be written compactly as
\begin{equation}
\small
\theta(s,t)
=
\boldsymbol{\phi}^{\mathrm T}(s)
\Delta\boldsymbol{L}(t).
\label{eq:theta_actuation_vector}
\end{equation}
Because the geometric and material parameters are time invariant, differentiating~\eqref{eq:theta_actuation_vector} with respect to time gives
\begin{equation}
\small
\dot{\theta}(s,t)
=
\boldsymbol{\phi}^{\mathrm T}(s)
\Delta\dot{\boldsymbol{L}}(t).
\label{eq:theta_velocity_mapping}
\end{equation}
Differentiating the backbone geometry~\eqref{eq:backbone_geometry} with respect to time further yields
\begin{equation}
\small
\dot{\boldsymbol{p}}(s,t)
=
\int_{0}^{s}
\begin{bmatrix}
-\sin\theta(\xi,t)\\
\cos\theta(\xi,t)
\end{bmatrix}
\dot{\theta}(\xi,t)
\,\mathrm{d}\xi .
\label{eq:backbone_velocity}
\end{equation}
Substituting~\eqref{eq:theta_velocity_mapping} into~\eqref{eq:backbone_velocity} gives
\begin{equation}
\small
\dot{\boldsymbol{p}}(s,t)
=
\boldsymbol{J}(s,t)
\Delta\dot{\boldsymbol{L}}(t),
\label{eq:differential_kinematics}
\end{equation}
where
\begin{equation}
\small
\boldsymbol{J}(s,t)
=
\int_{0}^{s}
\begin{bmatrix}
-\sin\theta(\xi,t)\\
\cos\theta(\xi,t)
\end{bmatrix}
\boldsymbol{\phi}^{\mathrm T}(\xi)
\,\mathrm{d}\xi
\in\mathbb{R}^{2\times n}
\label{eq:point_jacobian}
\end{equation}
is the point Jacobian at arc-length position $s$ with respect to the tendon-length variations. This matrix directly maps the multi-segmen actuation rates to the Cartesian velocity of any backbone position.

For the robot tip $s=L$, define $\boldsymbol{p}_{e}(t)
=
\boldsymbol{p}(L,t),
\boldsymbol{J}_{e}(t)
=
\boldsymbol{J}(L,t),
$
The tip differential kinematics can then be written as
\begin{equation}
\small
\dot{\boldsymbol{p}}_{e}(t)
=
\boldsymbol{J}_{e}(t)
\Delta\dot{\boldsymbol{L}}(t).
\label{eq:end_effector_differential_kinematics}
\end{equation}

The tip forward kinematics can be written as the following implicit nonlinear equation with respect to the actuation input and tip position:
\begin{equation} 
\small 
\mathcal{F}_{\mathrm{FK}} 
\left( 
\Delta\boldsymbol{L}(t), 
\boldsymbol{p}_{e}(t) 
\right) 
= 
\boldsymbol{0}, 
\label{eq:implicit_forward_kinematics} 
\end{equation} 
where $\mathcal{F}_{\mathrm{FK}}(\cdot)$ denotes the variable-curvature forward-kinematic residual at the tip, as jointly defined by~\eqref{eq:theta_length_global} and~\eqref{eq:backbone_geometry}. Specifically,
\begin{equation*} 
\small 
\mathcal{F}_{\mathrm{FK}} 
\left( 
\Delta\boldsymbol{L}(t), 
\boldsymbol{p}_{e}(t) 
\right) 
= 
\boldsymbol{p}_{e}(t) 
- 
\int_{0}^{L} 
\begin{bmatrix} 
\cos\theta(\xi,t)\\ 
\sin\theta(\xi,t) 
\end{bmatrix} 
\mathrm{d}\xi . 
\end{equation*} 
 
Given a desired tip trajectory 
$\boldsymbol{p}_{d}(t)\in\mathbb{R}^{2}$, 
the inverse kinematic problem amounts to determining the actuation input $\Delta\boldsymbol{L}(t)$ such that
\begin{equation} 
\small 
\boldsymbol{e}_{p}(t) 
= 
\mathcal{F}_{\mathrm{FK}} 
\left( 
\Delta\boldsymbol{L}(t), 
\boldsymbol{p}_{d}(t) 
\right) 
= 
\boldsymbol{0}. 
\label{eq:kinematic_residual} 
\end{equation} 
where $\boldsymbol{e}_{p}(t)\in\mathbb{R}^{2}$ denotes the tip kinematic residual. Differentiating~\eqref{eq:kinematic_residual} with respect to time and applying the chain rule yields
\begin{equation} 
\small 
\frac{\partial\mathcal{F}_{\mathrm{FK}}} 
{\partial\Delta\boldsymbol{L}} 
\Delta\dot{\boldsymbol{L}}(t) 
+ 
\frac{\partial\mathcal{F}_{\mathrm{FK}}} 
{\partial\boldsymbol{p}_{d}} 
\dot{\boldsymbol{p}}_{d}(t) 
= 
\dot{\boldsymbol{e}}_{p}(t). 
\label{eq:residual_chain_rule} 
\end{equation} 
From the tip forward kinematics,
\begin{equation*} 
\small 
\frac{\partial\mathcal{F}_{\mathrm{FK}}} 
{\partial\Delta\boldsymbol{L}} 
= 
-\boldsymbol{J}_{e}(t), 
\quad 
\frac{\partial\mathcal{F}_{\mathrm{FK}}} 
{\partial\boldsymbol{p}_{d}} 
= 
\boldsymbol{I}_{2}, 
\end{equation*} 
and~\eqref{eq:residual_chain_rule} reduces to
\begin{equation} 
\small 
\dot{\boldsymbol{e}}_{p}(t) 
= 
\dot{\boldsymbol{p}}_{d}(t) 
- 
\boldsymbol{J}_{e}(t) 
\Delta\dot{\boldsymbol{L}}(t). 
\label{eq:residual_dynamics} 
\end{equation} 
 
To ensure asymptotic convergence of the tip kinematic residual, the following first-order residual dynamics are imposed:
\begin{equation} 
\small 
\dot{\boldsymbol{e}}_{p}(t) 
+ 
k\boldsymbol{e}_{p}(t) 
= 
\boldsymbol{0}, 
\label{eq:residual_convergence_law} 
\end{equation} 
where $k>0$ denotes the residual-convergence gain. The resulting first-order dynamics provide smooth residual regulation while avoiding excessively aggressive actuation near the equilibrium. Substituting~\eqref{eq:residual_dynamics} into~\eqref{eq:residual_convergence_law} yields
\begin{equation} 
\small 
\boldsymbol{J}_{e}(t) 
\Delta\dot{\boldsymbol{L}}(t) 
= 
\dot{\boldsymbol{p}}_{d}(t) 
+ 
k\boldsymbol{e}_{p}(t). 
\label{eq:inverse_residual_equation} 
\end{equation} 
 
Since 
$\boldsymbol{J}_{e}(t)\in\mathbb{R}^{2\times n}$, 
the independently actuated multi-segment robot is kinematically redundant with respect to the two-dimensional tip task when $n>2$. Accordingly, a nominal actuation velocity can be obtained as
\begin{equation} 
\small 
\Delta\dot{\boldsymbol{L}}_{\mathrm{nom}}(t) 
= 
\boldsymbol{J}_{e}^{\dagger}(t) 
\left[ 
\dot{\boldsymbol{p}}_{d}(t) 
+ 
k\boldsymbol{e}_{p}(t) 
\right], 
\label{eq:nominal_inverse_kinematics} 
\end{equation} 
where $\boldsymbol{J}_{e}^{\dagger}(t)$ denotes the Moore--Penrose pseudoinverse of the tip Jacobian. 
 
When the robot approaches a kinematic singularity, numerical robustness can be improved by employing the damped least-squares pseudoinverse
\begin{equation} 
\small 
\boldsymbol{J}_{e}^{\#}(t) 
= 
\boldsymbol{J}_{e}^{\mathrm T}(t) 
\left[ 
\boldsymbol{J}_{e}(t) 
\boldsymbol{J}_{e}^{\mathrm T}(t) 
+ 
\mu^{2}\boldsymbol{I}_{2} 
\right]^{-1}, 
\label{eq:damped_pseudoinverse} 
\end{equation} 
where $\mu>0$ is the damping coefficient. The corresponding nominal actuation velocity is then given by
\begin{equation} 
\small 
\Delta\dot{\boldsymbol{L}}_{\mathrm{nom}}(t) 
= 
\boldsymbol{J}_{e}^{\#}(t) 
\left[ 
\dot{\boldsymbol{p}}_{d}(t) 
+ 
k\boldsymbol{e}_{p}(t) 
\right]. 
\label{eq:damped_nominal_inverse_kinematics} 
\end{equation} 
Since $k>0$, Eq.~\eqref{eq:residual_convergence_law} implies that the tip kinematic residual $\boldsymbol{e}_{p}(t)$ converges asymptotically to zero. Consequently, the robot tip asymptotically tracks the desired trajectory, completing the corresponding inverse kinematic solution.

\section{Real-Time Whole-Body Collision-Avoidance Framework}

Building on the multi-segmen continuum-robot kinematics, this section develops a real-time whole-body safety-control framework. Unlike methods that impose collision-avoidance constraints only at the robot tip, the proposed framework selects multiple safety-monitoring positions along the flexible backbone and constructs control barrier function constraints using their actuation-space point Jacobians. Tip trajectory tracking and whole-body collision avoidance are thereby unified in an actuation-space quadratic program.

\subsection{Actuation-Space System and Whole-Body Safe Set}

Taking the tendon-length variations as the actuation-space state of the robot configuration, define
\begin{equation*}
\small
\boldsymbol{q}(t)
=
\Delta\boldsymbol{L}(t)
\in\mathbb{R}^{n},
\quad
\boldsymbol{u}(t)
=
\Delta\dot{\boldsymbol{L}}(t)
\in\mathbb{R}^{n},
\label{eq:control_state_input}
\end{equation*}
where $\boldsymbol{q}(t)$ is the actuation state of the multi-segmen continuum robot and $\boldsymbol{u}(t)$ is the corresponding actuation-velocity input. The actuation-space system is therefore
\begin{equation}
\small
\dot{\boldsymbol{q}}(t)
=
\boldsymbol{u}(t).
\label{eq:actuation_space_system}
\end{equation}

From the forward kinematics, any backbone position at arc length $s\in[0,L]$ can be expressed as a function of the actuation state $\boldsymbol{q}(t)$, i.e.,
\begin{equation*}
\small
\boldsymbol{p}(s,t)
=
\boldsymbol{p}\bigl(s,\boldsymbol{q}(t)\bigr).
\end{equation*}
Accordingly,~\eqref{eq:differential_kinematics} gives
\begin{equation}
\small
\dot{\boldsymbol{p}}(s,t)
=
\boldsymbol{J}(s,t)\boldsymbol{u}(t),
\label{eq:whole_body_point_velocity}
\end{equation}

To characterize the collision safety of the continuous flexible backbone, select $N_\alpha$ safety-monitoring positions along the backbone
\begin{equation*}
\small
0<s_1<s_2<\cdots<s_{N_\alpha}\leq L,
\end{equation*}
and define the position and Jacobian of the $k$th safety point as
\begin{equation}
\small
\boldsymbol{p}_k(t)
=
\boldsymbol{p}(s_k,t),
\quad
\boldsymbol{J}_k(t)
=
\boldsymbol{J}(s_k,t),
\quad
k=1,\ldots,N_\alpha.
\label{eq:safety_point_definition}
\end{equation}
The Cartesian velocity of each safety point then satisfies
\begin{equation}
\small
\dot{\boldsymbol{p}}_k(t)
=
\boldsymbol{J}_k(t)\boldsymbol{u}(t).
\label{eq:safety_point_velocity}
\end{equation}

Consider an environment containing $N_o$ obstacles, where the center of the $o$th obstacle is denoted by
$\boldsymbol{p}_{o}(t)\in\mathbb{R}^{2}$. For the $k$th safety point and the $o$th obstacle, define the safety function
\begin{equation}
\small
h_{k,o}(\boldsymbol{q},t)
=
\left\|
\boldsymbol{p}_k(\boldsymbol{q})
-
\boldsymbol{p}_{o}(t)
\right\|^{2}
-
d_{o}^{2},
\label{eq:whole_body_barrier_function}
\end{equation}
where $d_o>0$ is the prescribed minimum clearance from the $o$th obstacle. The safe set of the $k$th safety point relative to the $o$th obstacle is then defined as
\begin{equation*}
\small
\mathcal{C}_{k,o}(t)
=
\left\{
\boldsymbol{q}\in\mathbb{R}^{n}
\mid
h_{k,o}(\boldsymbol{q},t)\geq0
\right\},
\end{equation*}
The whole-body safe set is
\begin{equation}
\small
\mathcal{C}(t)
=
\bigcap_{k=1}^{N_\alpha}
\bigcap_{o=1}^{N_o}
\mathcal{C}_{k,o}(t).
\label{eq:whole_body_safe_set}
\end{equation}
Therefore, when $\boldsymbol{q}(t)\in\mathcal{C}(t)$, all safety-monitoring positions maintain the prescribed minimum clearance from the environmental obstacles.
\subsection{Multipoint Control Barrier Function Constraints}

To ensure forward invariance of the safe set~\eqref{eq:whole_body_safe_set}, differentiating the safety function~\eqref{eq:whole_body_barrier_function} with respect to time gives
\begin{equation}
\small
\dot{h}_{k,o}
=
2
\left[
\boldsymbol{p}_k(t)-\boldsymbol{p}_{o}(t)
\right]^{\mathrm T}
\left[
\dot{\boldsymbol{p}}_k(t)-\dot{\boldsymbol{p}}_{o}(t)
\right].
\label{eq:barrier_derivative_position}
\end{equation}
where $\boldsymbol{p}_{o}(t)$ and $\dot{\boldsymbol{p}}_{o}(t)$ denote the position and velocity of the $o$th obstacle, respectively. For a stationary obstacle, $\dot{\boldsymbol{p}}_{o}(t)=\boldsymbol{0}$.

Combining this result with~\eqref{eq:safety_point_velocity} gives
\begin{equation}
\small
\dot{h}_{k,o}
=
2
\left[
\boldsymbol{p}_k(t)-\boldsymbol{p}_{o}(t)
\right]^{\mathrm T}
\left[
\boldsymbol{J}_k(t)\boldsymbol{u}(t)
-
\dot{\boldsymbol{p}}_{o}(t)
\right].
\label{eq:barrier_derivative}
\end{equation}

For each safety-point--obstacle pair $(k,o)$, impose the first-order control barrier function condition
\begin{equation}
\small
\dot{h}_{k,o}
+
\gamma h_{k,o}
\geq0,
\quad
k=1,\ldots,N_\alpha,\quad
o=1,\ldots,N_o,
\label{eq:cbf_general_condition}
\end{equation}
where $\gamma>0$ is the control barrier function gain. Substituting~\eqref{eq:barrier_derivative} into~\eqref{eq:cbf_general_condition} yields the linear safety constraint on the actuation-velocity input $\boldsymbol{u}(t)$
\begin{equation}
\small
\begin{aligned}
&
2
\left[
\boldsymbol{p}_k(t)-\boldsymbol{p}_{o}(t)
\right]^{\mathrm T}
\boldsymbol{J}_k(t)\boldsymbol{u}(t)
\\
&\quad
-
2
\left[
\boldsymbol{p}_k(t)-\boldsymbol{p}_{o}(t)
\right]^{\mathrm T}
\dot{\boldsymbol{p}}_{o}(t)
+
\gamma h_{k,o}
\geq0 .
\end{aligned}
\label{eq:pointwise_cbf_constraint}
\end{equation}
Thus, the collision-avoidance condition between each safety-monitoring point and obstacle is converted into an affine inequality constraint on the actuation-velocity input. Enforcing all $N_\alpha N_o$ constraints maps the local safety requirements at multiple backbone positions into actuation space, yielding a unified constraint set for whole-body continuum-robot motion.

For subsequent optimization, rewrite~\eqref{eq:pointwise_cbf_constraint} as
\begin{equation*}
\small
\boldsymbol{A}_{k,o}(t)\boldsymbol{u}(t)
\geq
b_{k,o}(t),
\end{equation*}
where
\begin{equation*}
\small
\begin{aligned}
\boldsymbol{A}_{k,o}(t)
&=
2
\left[
\boldsymbol{p}_k(t)-\boldsymbol{p}_{o}(t)
\right]^{\mathrm T}
\boldsymbol{J}_k(t),
\\
b_{k,o}(t)
&=
2
\left[
\boldsymbol{p}_k(t)-\boldsymbol{p}_{o}(t)
\right]^{\mathrm T}
\dot{\boldsymbol{p}}_{o}(t)
-
\gamma h_{k,o}.
\end{aligned}
\end{equation*}
Row-wise stacking the constraints for all safety-point--obstacle pairs yields
\begin{equation}
\small
\boldsymbol{A}_{h}(t)\boldsymbol{u}(t)
\geq
\boldsymbol{b}_{h}(t),
\label{eq:stacked_cbf_constraint}
\end{equation}
where $\boldsymbol{A}_{h}(t)\in\mathbb{R}^{N_\alpha N_o\times n}$ and
$\boldsymbol{b}_{h}(t)\in\mathbb{R}^{N_\alpha N_o}$ are obtained by row-wise stacking
$\boldsymbol{A}_{k,o}(t)$ and $b_{k,o}(t)$, respectively, and represent the complete whole-body CBF constraints.

\subsection{Real-Time Whole-Body Safe Motion Control via CBF-QP}
The inverse kinematic model in the preceding section provides the nominal actuation velocity $\Delta\dot{\boldsymbol{L}}_{\mathrm{nom}}(t)$ for tip trajectory tracking. Since control is performed directly in the $n$-dimensional actuation space, the actual actuation velocity $\boldsymbol{u}(t)=\Delta\dot{\boldsymbol{L}}(t)\in\mathbb{R}^{n}$ is selected as the optimization variable. To minimally modify the nominal input while enforcing whole-body safety constraints, the following CBF quadratic program is solved online:
\begin{equation}
\small
\begin{aligned}
\boldsymbol{u}^{*}(t)
=
\arg\min_{\boldsymbol{u}(t)}
\quad&
\frac{1}{2}
\left\|
\boldsymbol{u}(t)
-
\Delta\dot{\boldsymbol{L}}_{\mathrm{nom}}(t)
\right\|^{2}
\\
\mathrm{s.t.}\quad&
\boldsymbol{A}_{h}(t)\boldsymbol{u}(t)
\geq
\boldsymbol{b}_{h}(t),
\\
&
\boldsymbol{u}_{\min}
\leq
\boldsymbol{u}(t)
\leq
\boldsymbol{u}_{\max}.
\end{aligned}
\label{eq:whole_body_cbf_qp}
\end{equation}
Here, $\boldsymbol{u}^{*}(t)$ is the optimal actuation velocity satisfying the whole-body safety constraints,
and $\boldsymbol{u}_{\min}$ and $\boldsymbol{u}_{\max}$ are the minimum and maximum allowable tendon velocities of each segmen, respectively.
When the safety constraints are feasible, the optimization minimally modifies the nominal inverse-kinematic input,
thereby balancing tip trajectory tracking and whole-body safety.

If the initial actuation state satisfies
$\boldsymbol{q}(0)\in\mathcal{C}(0)$,
and~\eqref{eq:whole_body_cbf_qp} remains feasible during subsequent motion,
the control barrier function constraints guarantee forward invariance of the time-varying safe set $\mathcal{C}(t)$, i.e.,
\begin{equation}
\small
\boldsymbol{q}(t)\in\mathcal{C}(t),
\quad
\forall t\geq0.
\label{eq:whole_body_safety_guarantee}
\end{equation}
Therefore, all selected backbone safety-monitoring positions maintain the prescribed clearance constraints throughout the motion.

For clarity, the online implementation of the proposed control law
integrates the inverse-kinematic computation, whole-body CBF construction, and QP solution into the real-time procedure summarized in Algorithm~\ref{alg:whole_body_safe_control}.
\begin{algorithm}[h]
    \caption{Real-Time Whole-Body Safe Motion Control}
    \footnotesize
    \label{alg:whole_body_safe_control}
    \begin{algorithmic}[1]
        \STATE \textbf{Inputs:} Desired trajectory $\boldsymbol{p}_d(t)$, initial actuation state $\boldsymbol{q}_0$, safety points $\{s_k\}_{k=1}^{N_\alpha}$, and obstacle information $\{\boldsymbol{p}_o(t),\dot{\boldsymbol{p}}_o(t)\}_{o=1}^{N_o}$.
        \STATE \textbf{Initialization:} Set $\boldsymbol{q}(0)=\boldsymbol{q}_0$ and compute $\Xi_i(s)$ and $\Gamma_i$, $i=1,\ldots,n$.
        \FOR{$\ell = 0$ to $N_t-1$}
            \STATE \textbf{Kinematics:} Evaluate $\boldsymbol{p}(s,t_\ell)$ and $\boldsymbol{J}(s,t_\ell)$ from the current actuation state $\boldsymbol{q}(t_\ell)$.
            \STATE \textbf{Nominal input:} Compute $\boldsymbol{J}_e(t_\ell)$ and obtain $\Delta\dot{\boldsymbol{L}}_{\mathrm{nom}}(t_\ell)$ from Eq.~\eqref{eq:nominal_inverse_kinematics}.
            \STATE \textbf{Safety update:} Evaluate $\boldsymbol{p}_k(t_\ell)$, $\boldsymbol{J}_k(t_\ell)$, and $h_{k,o}$ for $k=1,\ldots,N_\alpha$ and $o=1,\ldots,N_o$.
            \STATE \textbf{CBF constraints:} Construct $\boldsymbol{A}_{h}(t_\ell)$ and $\boldsymbol{b}_{h}(t_\ell)$ according to Eq.~\eqref{eq:stacked_cbf_constraint}.
            \STATE \textbf{QP solve:} Solve Eq.~\eqref{eq:whole_body_cbf_qp} to obtain $\boldsymbol{u}^{*}(t_\ell)$.
            \STATE \textbf{State update:} $\boldsymbol{q}(t_{\ell+1}) = \boldsymbol{q}(t_\ell) + \boldsymbol{u}^{*}(t_\ell)\Delta t$.
        \ENDFOR
    \end{algorithmic}
\end{algorithm}
\begin{table}[!t]
\centering
\makeatletter
\long\def\@makecaption#1#2{%
  \begin{center}
    {\footnotesize #1: {\scshape #2}\par}
  \end{center}
  \vskip\abovecaptionskip\relax
}
\makeatother
\caption{Simulation Parameters}
\label{tab:simulation-parameters}
\footnotesize
\setlength{\tabcolsep}{2pt}
\renewcommand{\arraystretch}{1.05}
\begin{tabular}{
@{}
p{0.32\columnwidth}
p{0.14\columnwidth}
p{0.39\columnwidth}
p{0.10\columnwidth}
@{}}
\toprule
\textbf{Parameter}
& \textbf{Symbol}
& \textbf{Value or profile}
& \textbf{Unit} \\
\midrule

Number of segments
& $N$
& $3$
& -- \\

Total backbone length
& $L$
& $0.30$
& m \\

Segment lengths
& $L_i$
& $[0.12,\,0.10,\,0.08]$
& m \\

Young's moduli
& $E_i$
& $[6.0,\,7.0,\,5.5]$
& MPa \\


Diameter profile
& $D(s)$
& $6.4\times10^{-3}(1-0.5s/L)$
& m \\

Second moment of area
& $I(s)$
& $\pi D^4(s)/64$
& $\mathrm{m^4}$ \\


Tendon-spacing profile
& $W(s)$
& $0.020(1-0.75(s/L)^3)$
& m \\








\bottomrule
\end{tabular}

\end{table}
\section{Numerical Validation}
Numerical simulations evaluate the kinematic accuracy, whole-body collision avoidance, and computational efficiency of the proposed framework. Kinematic predictions are compared with GVS results~\cite{mathew2025reduced}. 
Whole-body collision avoidance is then evaluated with static obstacles and during trajectory tracking with moving obstacles.

\begin{figure*}[t]
\centering
\captionsetup[subfloat]{font=scriptsize}
\subfloat[\scriptsize Robot shape comparison
\label{fig:shape-case-1}]
{\includegraphics[width=0.23\textwidth]
{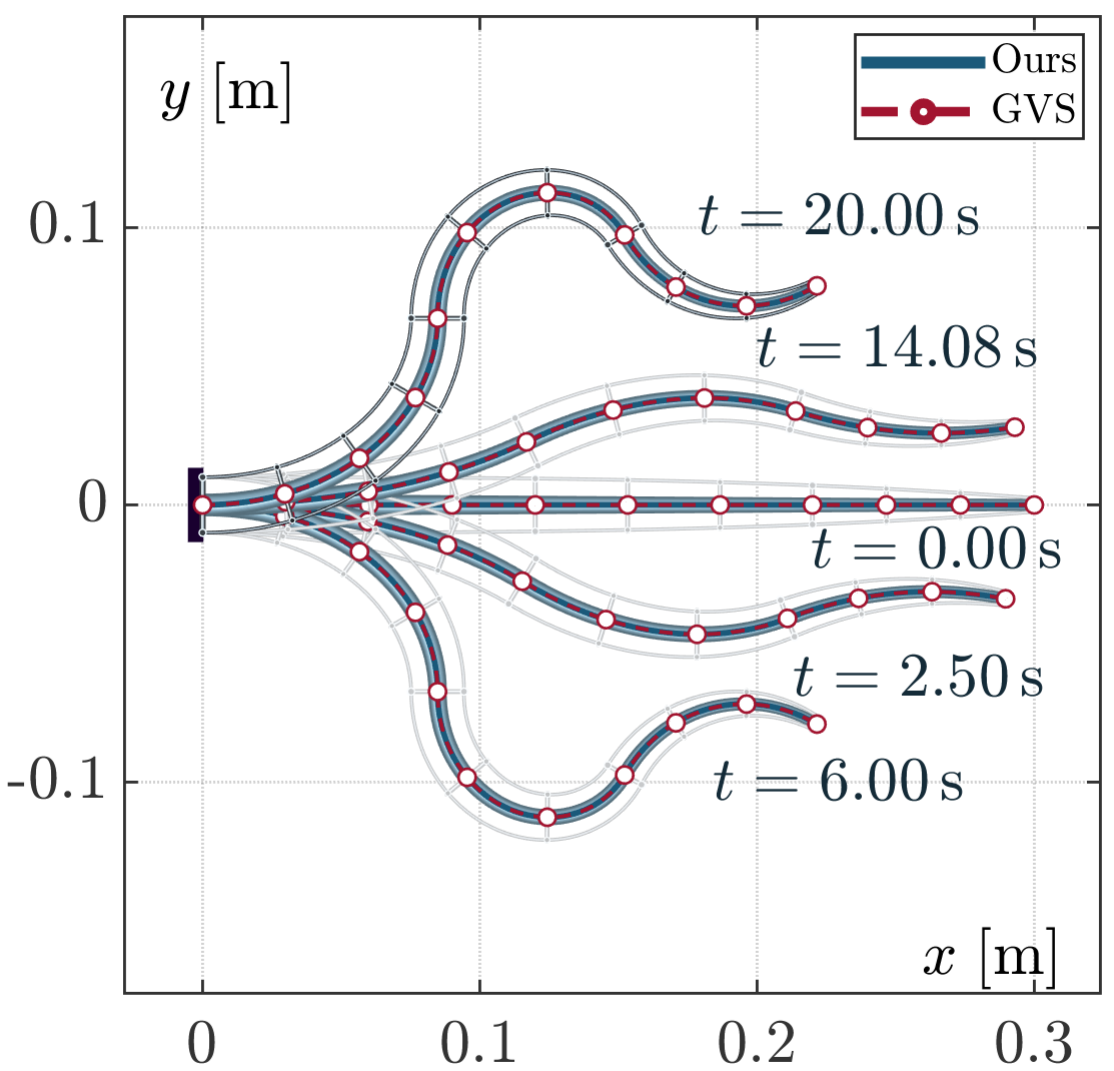}}
\hfill
\subfloat[\scriptsize $S_1$ curvature error
\label{fig:shape-case-2}]
{\includegraphics[width=0.25\textwidth]
{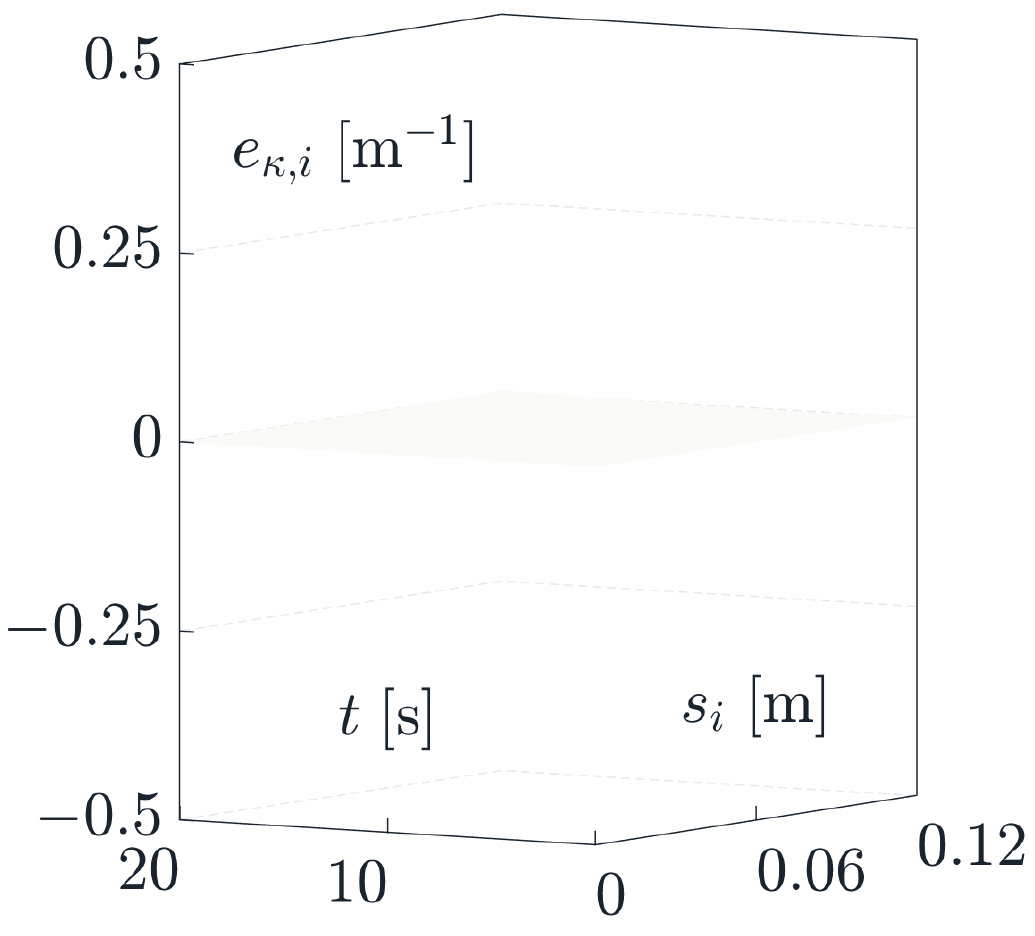}}
\hfill
\subfloat[\scriptsize $S_2$ curvature error
\label{fig:shape-case-3}]
{\includegraphics[width=0.25\textwidth]
{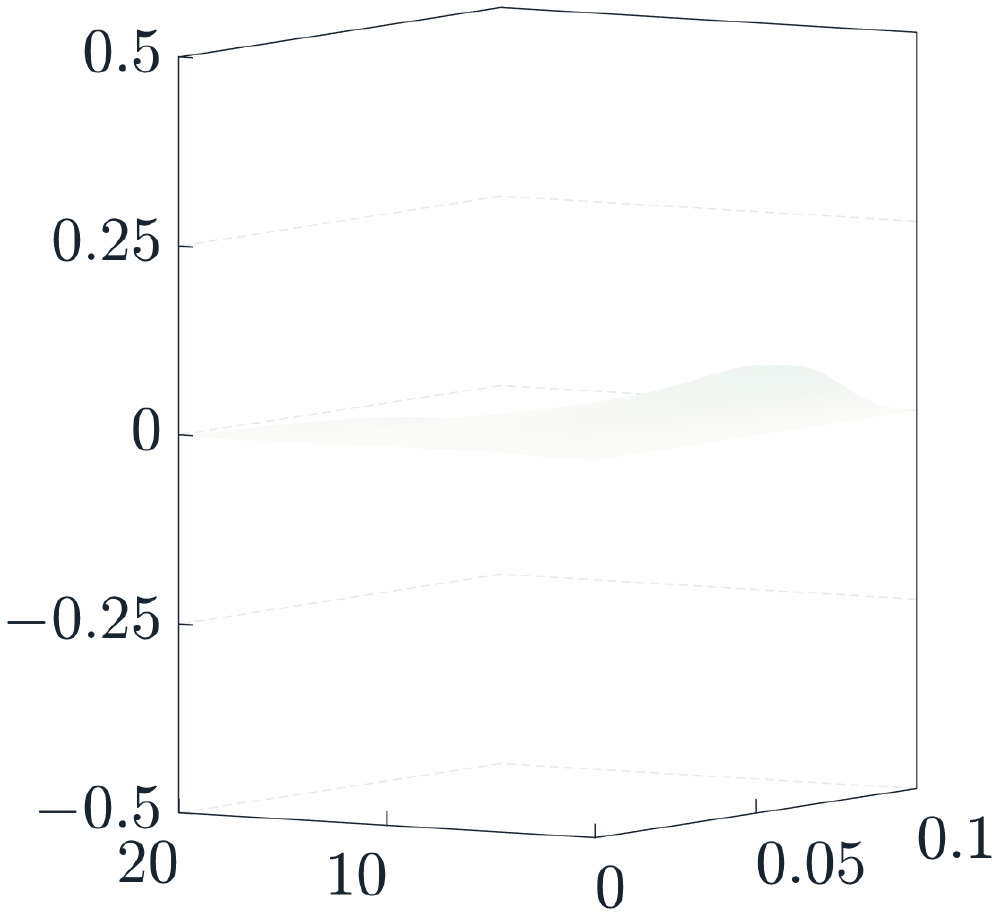}}
\hfill
\subfloat[\scriptsize $S_3$ curvature error
\label{fig:shape-case-4}]
{\includegraphics[width=0.27\textwidth]
{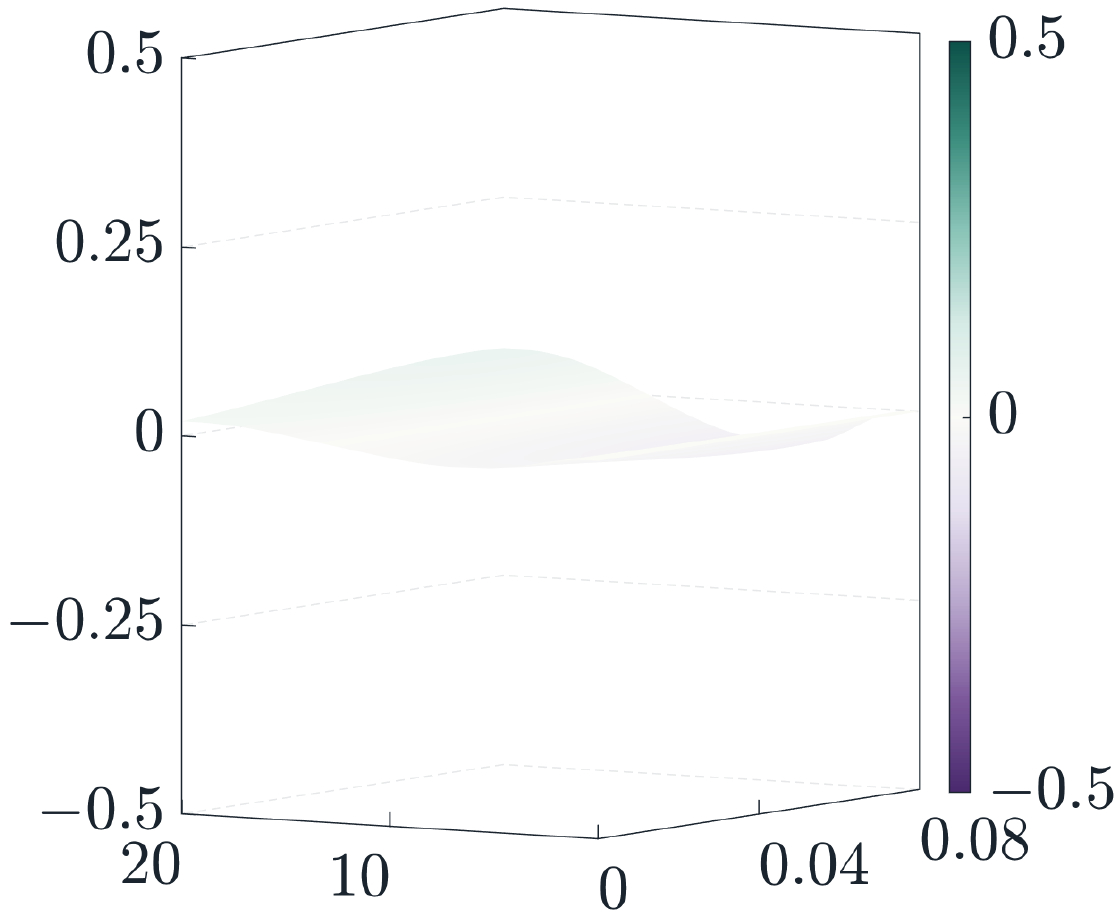}}
\caption{Kinematic validation against GVS: shape comparison and segment-wise curvature errors.}
\label{fig:sim-mod}
\end{figure*}

\begin{figure*}[t]
\centering
\captionsetup[subfloat]{font=scriptsize}
\subfloat[\scriptsize Static case 1 with CBF
\label{fig:shape-case-1}]
{\includegraphics[width=0.24\textwidth]
{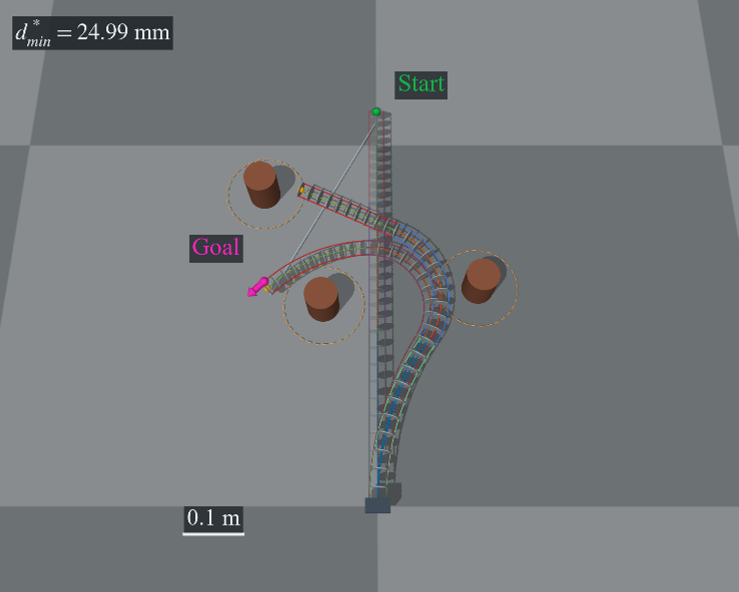}}
\hfill
\subfloat[\scriptsize Static case 1 without CBF
\label{fig:shape-case-2}]
{\includegraphics[width=0.24\textwidth]
{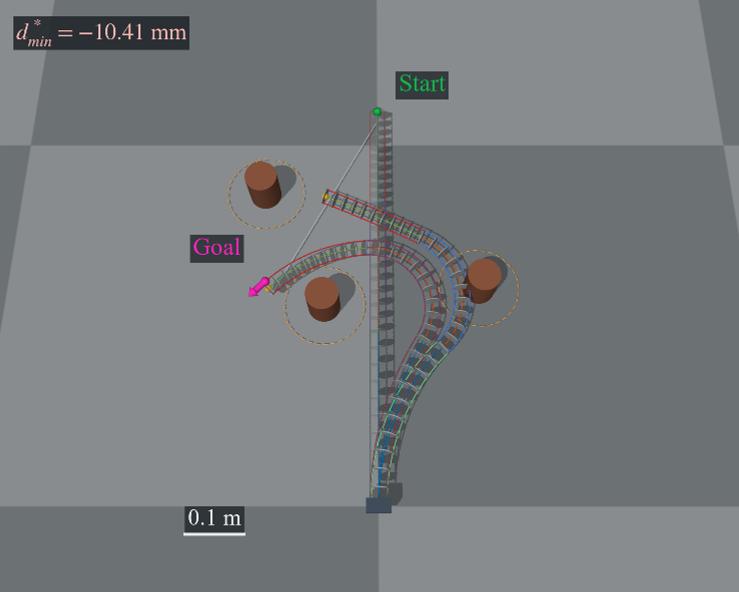}}
\hfill
\subfloat[\scriptsize Static case 2 with CBF
\label{fig:shape-case-3}]
{\includegraphics[width=0.24\textwidth]
{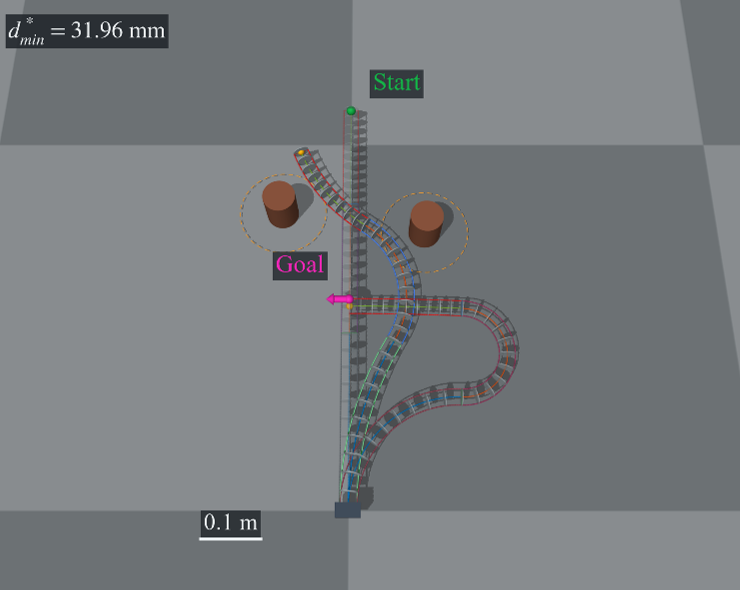}}
\hfill
\subfloat[\scriptsize Static case 2 without CBF
\label{fig:shape-case-4}]
{\includegraphics[width=0.24\textwidth]
{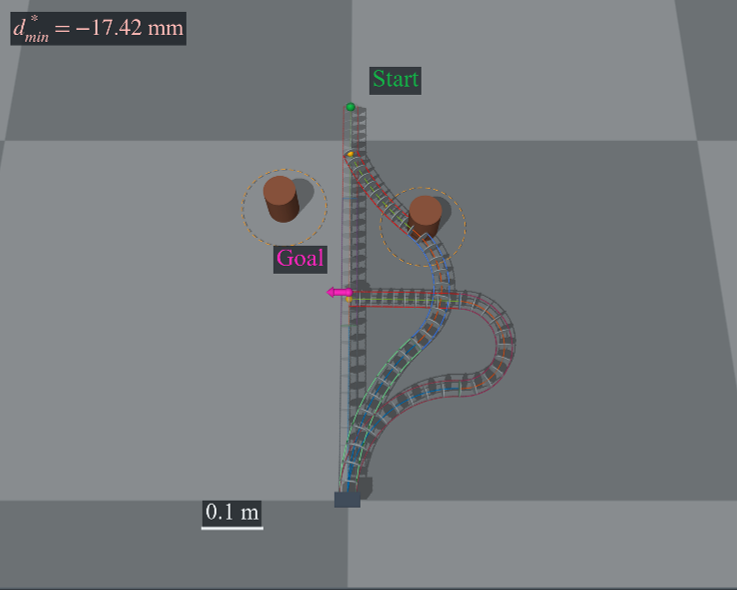}}

\par\vspace{-1.8mm}

\subfloat[\scriptsize Dynamic case with CBF
\label{fig:shape-case-1}]
{\includegraphics[width=0.24\textwidth]
{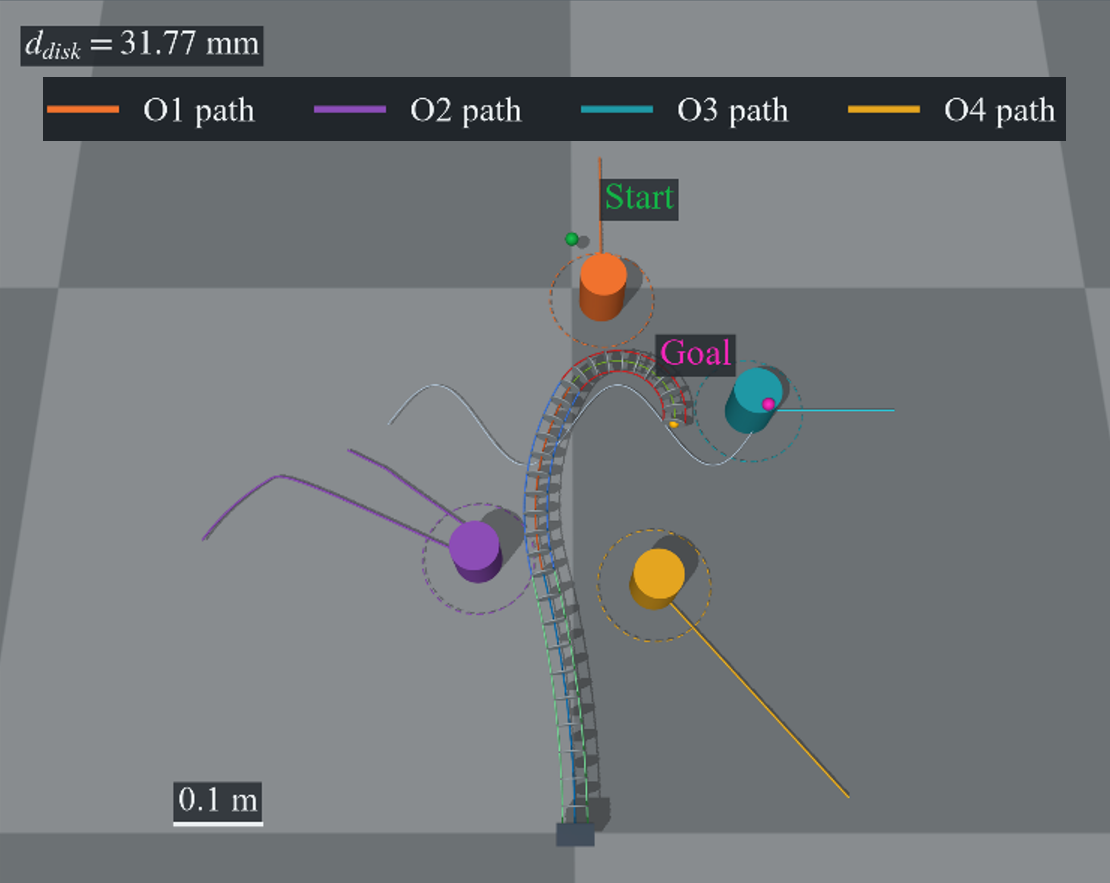}}
\hfill
\subfloat[\scriptsize Dynamic case without CBF
\label{fig:shape-case-2}]
{\includegraphics[width=0.24\textwidth]
{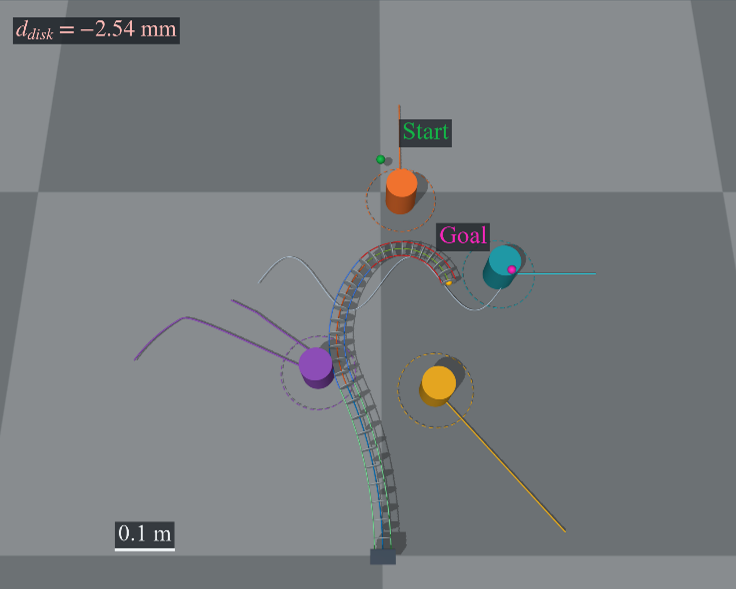}}
\hfill
\subfloat[\scriptsize Hole traversal with CBF
\label{fig:shape-case-3}]
{\includegraphics[width=0.24\textwidth]
{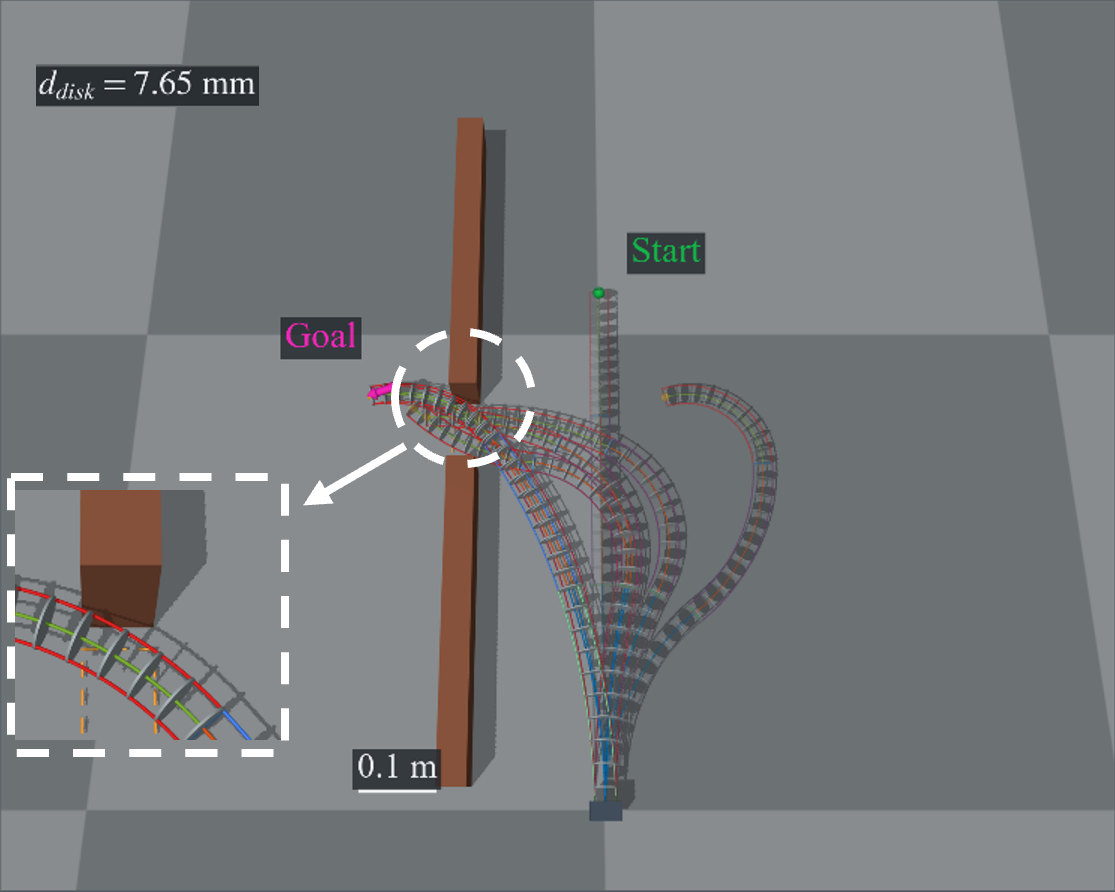}}
\hfill
\subfloat[\scriptsize Hole traversal without CBF
\label{fig:shape-case-4}]
{\includegraphics[width=0.24\textwidth]
{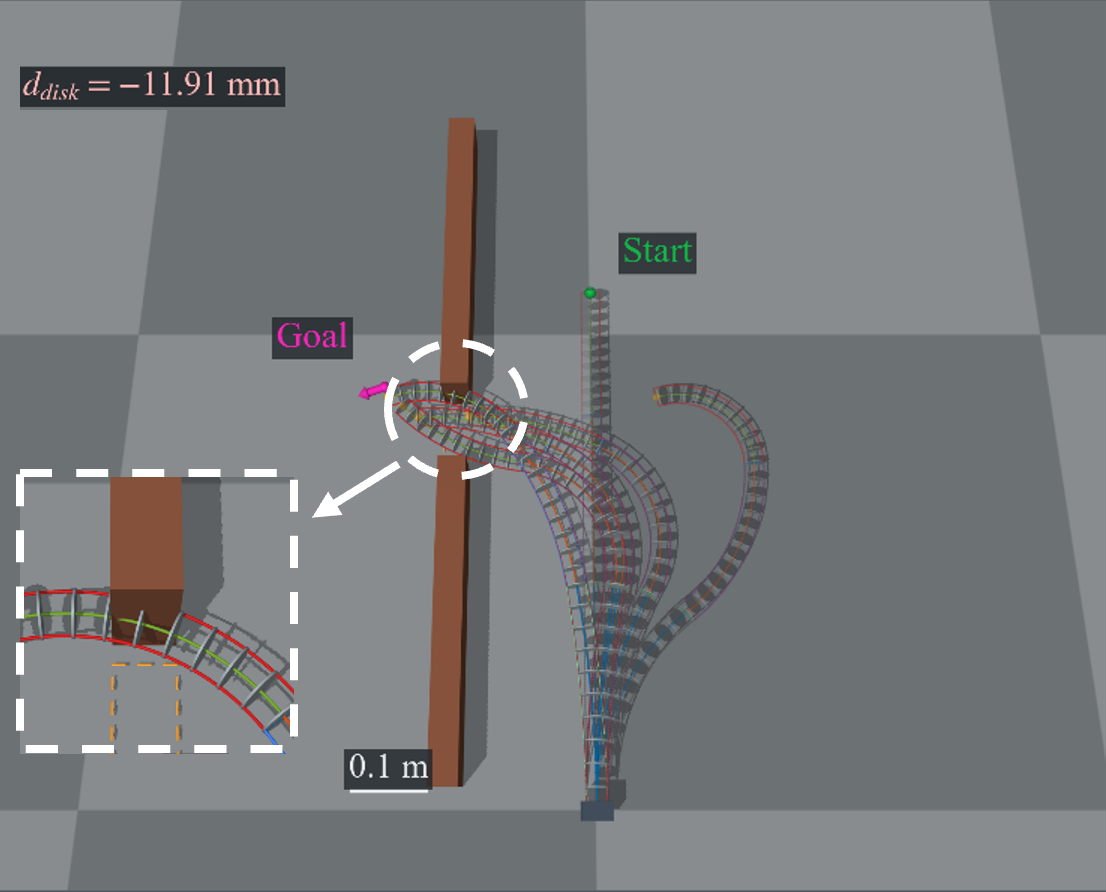}}


\caption{Sim-to-sim validation of real-time whole-body safe motion generation in MuJoCo: static and dynamic obstacle avoidance and hole traversal.}
\label{fig:sim2sim}
\end{figure*}

\subsection{Simulation Setup}

Table~\ref{tab:simulation-parameters} summarizes the parameters used for the kinematic validation in Fig.~\ref{fig:sim-mod}. The proposed model and GVS use identical geometric and material properties and tendon-length inputs. For the control tests, the segment lengths and constant tendon spacings are set to $L_i=[0.30,0.25,0.20]_i\,\mathrm{m}$ and $W_i=[16.5,23.5,30.5]_i\,\mathrm{mm}$, respectively, with $i\in\{1,2,3\}$, giving a total length of $L=0.75\,\mathrm{m}$.

Additional sim-to-sim validation is conducted in MuJoCo to assess controller performance in a physics-based environment, covering whole-body collision avoidance with static and moving obstacles and hole traversal. Comparisons are conducted with the hard CBF constraints enabled and disabled under identical task settings.

\subsection{Numerical Results and Analysis}

\subsubsection{Variable-Curvature Model Accuracy}
Fig.~\ref{fig:sim-mod}(a) shows close agreement between the predicted backbone shapes and the GVS solutions. Over the 20-s trajectory, the maximum absolute curvature errors are
\(2.109\times10^{-3}\), \(4.929\times10^{-2}\), and
\(5.223\times10^{-2}\,\mathrm{m}^{-1}\) for \(S_1\), \(S_2\), and \(S_3\), respectively, confirming accurate representation of the tested nonuniform bending behavior.


\subsubsection{Whole-Body Collision Avoidance}
Figs.~\ref{fig:sim2sim}(a)--(d) show that the proposed CBF-constrained
controller maintains whole-body clearance while reaching the targets in the
static scenarios. Across 100 independent MuJoCo trials per scenario, it
achieved a \textbf{96\% collision-free success rate}; the few failures were
mainly due to conservative CBF constraints causing the robot to become
temporarily trapped. In comparison, the non-CBF baseline achieved only about
\(60\%\) success, mainly because obstacle contact could lock the robot into
unfavorable postures. In the dynamic-obstacle case
(Figs.~\ref{fig:sim2sim}(e)--(f)), the proposed method achieved a
\textbf{100\% collision-free success rate} over 100 trials, compared with
approximately \(80\%\) for the non-CBF baseline, whose higher success rate
than in the static case was partly due to the obstacle moving away from the
robot during the motion.

\subsubsection{Computation Time Evaluation}
\begin{table}[!t]
\centering
\caption{CBF Sampling Density and Computational Cost}
\label{tab:cbf-sampling-runtime}
\footnotesize
\setlength{\tabcolsep}{3pt}
\renewcommand{\arraystretch}{1.15}

\begin{tabular*}{\columnwidth}{
    @{\extracolsep{\fill}}ccc@{}
}
\toprule
\makecell{\textbf{CBF sampling}\\\textbf{points}}
& $\boldsymbol{\Delta s}\,[\mathrm{m}]$
& \makecell{\textbf{Mean control-step}\\\textbf{time (ms)}} \\
\midrule

100 & 0.00758 & 1.1692 \\
200 & 0.00377 & 1.9870 \\
300 & 0.00251 & 2.9671 \\
400 & 0.00188 & 3.6831 \\
500 & 0.00150 & 5.0753 \\
600 & 0.00125 & 6.6641 \\

\bottomrule
\end{tabular*}
\end{table}
Table~\ref{tab:cbf-sampling-runtime} shows that denser backbone sampling increases computational cost by introducing more CBF constraints. Nevertheless, the compact kinematic formulation and actuation-space QP retain millisecond-level average computation times, with the number of optimization variables independent of the sampling density. Even with 600 monitoring points spaced approximately $1.25\,\mathrm{mm}$ apart, the mean control-step time is only $6.66\,\mathrm{\textbf{ms}}$. These results support the practical feasibility of real-time whole-body collision avoidance with dense, finite backbone sampling.

\section{Conclusion}
\label{sec:conclusion}
This paper presented a unified actuation-space framework for real-time whole-body safe motion generation of planar multi-segment tendon-driven continuum robots. The variable-curvature model captures nonuniform tendon spacing and bending stiffness and provides analytical point Jacobians for multipoint CBF-QP control. Numerical results show close agreement with GVS and demonstrate effective whole-body safety in static, dynamic, and constrained-passage scenarios. Over 100 MuJoCo trials, the proposed method achieved collision-free success rates of 96\% and 100\% in the static and dynamic scenarios, respectively, compared with approximately 60\% and 80\% for the non-CBF baseline. With 600 monitoring points, the mean control-step time is $6.66\,\mathrm{ms}$, supporting the computational feasibility of dense whole-body safety monitoring. Future work will focus on hardware validation and improving task completion in tightly constrained passages.
\bibliographystyle{IEEEtran}
\bibliography{reference}

\begin{thebibliography}{10}
\providecommand{\url}[1]{#1}
\csname url@samestyle\endcsname
\providecommand{\newblock}{\relax}
\providecommand{\bibinfo}[2]{#2}
\providecommand{\BIBentrySTDinterwordspacing}{\spaceskip=0pt\relax}
\providecommand{\BIBentryALTinterwordstretchfactor}{4}
\providecommand{\BIBentryALTinterwordspacing}{\spaceskip=\fontdimen2\font plus
\BIBentryALTinterwordstretchfactor\fontdimen3\font minus \fontdimen4\font\relax}
\providecommand{\BIBforeignlanguage}[2]{{%
\expandafter\ifx\csname l@#1\endcsname\relax
\typeout{** WARNING: IEEEtran.bst: No hyphenation pattern has been}%
\typeout{** loaded for the language `#1'. Using the pattern for}%
\typeout{** the default language instead.}%
\else
\language=\csname l@#1\endcsname
\fi
#2}}
\providecommand{\BIBdecl}{\relax}
\BIBdecl

\bibitem{burgner2015continuum}
J.~Burgner-Kahrs, D.~C. Rucker, and H.~Choset, ``Continuum robots for medical applications: A survey,'' \emph{IEEE Transactions on Robotics}, vol.~31, no.~6, pp. 1261--1280, 2015.

\bibitem{russo2023continuum}
M.~Russo, S.~M.~H. Sadati, X.~Dong, A.~Mohammad, I.~D. Walker, C.~Bergeles, K.~Xu, and D.~A. Axinte, ``Continuum robots: An overview,'' \emph{Advanced Intelligent Systems}, vol.~5, no.~5, p. 2200367, 2023.

\bibitem{rao2021modeling}
P.~Rao, Q.~Peyron, S.~Lilge, and J.~Burgner-Kahrs, ``How to model tendon-driven continuum robots and benchmark modelling performance,'' \emph{Frontiers in Robotics and AI}, vol.~7, p. 630245, 2021.

\bibitem{rucker2011statics}
D.~C. Rucker and R.~J. Webster, ``Statics and dynamics of continuum robots with general tendon routing and external loading,'' \emph{IEEE Transactions on Robotics}, vol.~27, no.~6, pp. 1033--1044, 2011.

\bibitem{mahl2014variable}
T.~Mahl, A.~Hildebrandt, and O.~Sawodny, ``A variable curvature continuum kinematics for kinematic control of the bionic handling assistant,'' \emph{IEEE Transactions on Robotics}, vol.~30, no.~4, pp. 935--949, 2014.

\bibitem{renda2020geometric}
F.~Renda, C.~Armanini, V.~Lebastard, F.~Candelier, and F.~Boyer, ``A geometric variable-strain approach for static modeling of soft manipulators with tendon and fluidic actuation,'' \emph{IEEE Robotics and Automation Letters}, vol.~5, no.~3, pp. 4006--4013, 2020.

\bibitem{ataka2016safer}
A.~Ataka, P.~Qi, A.~Shiva, A.~Shafti, H.~A. Wurdemann, P.~Dasgupta, and K.~Althoefer, ``Towards safer obstacle avoidance for continuum-style manipulator in dynamic environments,'' in \emph{2016 6th IEEE International Conference on Biomedical Robotics and Biomechatronics (BioRob)}.\hskip 1em plus 0.5em minus 0.4em\relax IEEE, 2016, pp. 600--605.

\bibitem{li2024active}
J.~Li, D.~Li, C.~Wang, W.~Guo, Z.~Wang, Z.~Zhang, and H.~Liu, ``Active collision avoidance for teleoperated multi-segment continuum robots toward minimally invasive surgery,'' \emph{The International Journal of Robotics Research}, vol.~43, no.~7, pp. 918--941, 2024.

\bibitem{jones2006kinematics}
B.~A. Jones and I.~D. Walker, ``Kinematics for multisection continuum robots,'' \emph{IEEE Transactions on Robotics}, vol.~22, no.~1, pp. 43--55, 2006.

\bibitem{webster2010design}
R.~J. Webster and B.~A. Jones, ``Design and kinematic modeling of constant curvature continuum robots: A review,'' \emph{The International Journal of Robotics Research}, vol.~29, no.~13, pp. 1661--1683, 2010.

\bibitem{camarillo2008mechanics}
D.~B. Camarillo, C.~F. Milne, C.~R. Carlson, M.~R. Zinn, and J.~K. Salisbury, ``Mechanics modeling of tendon-driven continuum manipulators,'' \emph{IEEE Transactions on Robotics}, vol.~24, no.~6, pp. 1262--1273, 2008.

\bibitem{janabi2021cosserat}
F.~Janabi-Sharifi, A.~Jalali, and I.~D. Walker, ``{Cosserat} rod-based dynamic modeling of tendon-driven continuum robots: A tutorial,'' \emph{IEEE Access}, vol.~9, pp. 68\,703--68\,719, 2021.

\bibitem{till2019realtime}
J.~Till, V.~Aloi, and C.~Rucker, ``Real-time dynamics of soft and continuum robots based on {Cosserat} rod models,'' \emph{The International Journal of Robotics Research}, vol.~38, no.~6, pp. 723--746, 2019.

\bibitem{ba2024analytical}
W.~Ba, J.-C. Chang, J.~Liu, X.~Wang, X.~Dong, and D.~Axinte, ``An analytical differential kinematics-based method for controlling tendon-driven continuum robots,'' \emph{Robotics and Autonomous Systems}, vol. 171, p. 104562, 2024.

\bibitem{Wu2026LASEM}
K.~Wu, Y.~Wang, K.~Henry, C.~Stefanini, and G.~Zheng, ``Lightweight kinematic and static modeling of cable-driven continuum robots via actuation-space energy formulation,'' \emph{IEEE Robotics and Automation Letters}, vol.~11, no.~5, pp. 5853--5860, 2026.

\bibitem{rao2024contact}
P.~Rao, O.~Salzman, and J.~Burgner-Kahrs, ``Towards contact-aided motion planning for tendon-driven continuum robots,'' \emph{IEEE Robotics and Automation Letters}, vol.~9, no.~5, pp. 4687--4694, 2024.

\bibitem{hachen2025nonlinear}
M.~Hachen, C.~Shentu, S.~Lilge, and J.~Burgner-Kahrs, ``A non-linear model predictive task-space controller satisfying shape constraints for tendon-driven continuum robots,'' \emph{IEEE Robotics and Automation Letters}, vol.~10, no.~3, pp. 2438--2445, 2025.

\bibitem{gao2024path}
G.~Gao, D.~Li, K.~Liu, Y.~Ge, and C.~Song, ``A study on path-planning algorithm for a multi-section continuum robot in confined multi-obstacle environments,'' \emph{Robotica}, vol.~42, no.~10, pp. 3324--3347, 2024.

\bibitem{ames2017control}
A.~D. Ames, X.~Xu, J.~W. Grizzle, and P.~Tabuada, ``Control barrier function based quadratic programs for safety critical systems,'' \emph{IEEE Transactions on Automatic Control}, vol.~62, no.~8, pp. 3861--3876, 2017.

\bibitem{wong2026closedform}
K.~Wong, M.~St{\"o}lzle, W.~Xiao, and D.~Rus, ``A closed-form {CLF-CBF} controller for whole-body continuum soft robot collision avoidance,'' in \emph{IEEE International Conference on Soft Robotics (RoboSoft)}, 2026, arXiv:2603.19424.

\bibitem{yang2026lightweightC}
F.~Yang, H.~Yang, I.~Alsarraj, Y.~Wang, and K.~Wu, ``Lightweight dynamic modeling of cable-driven continuum robots based on actuation-space energy formulation,'' \emph{IEEE Robotics and Automation Letters}, 2026.

\bibitem{mathew2025reduced}
A.~T. Mathew, D.~Feliu-Talegon, A.~Y. Alkayas, F.~Boyer, and F.~Renda, ``Reduced order modeling of hybrid soft-rigid robots using global, local, and state-dependent strain parameterization,'' \emph{The International Journal of Robotics Research}, vol.~44, no.~1, pp. 129--154, 2025.

\end{thebibliography}

\end{document}